\documentclass[journal]{IEEEtran}

\usepackage[T1]{fontenc}
\usepackage{times}
\usepackage{dcolumn}
\usepackage{endnotes}
\usepackage{graphics}
\usepackage{times}
\usepackage{epsfig}
\usepackage{graphicx}
\usepackage{textcomp}
\usepackage{amssymb,amsmath}
\usepackage{balance}

\usepackage{listings}
\usepackage[]{algorithm2e}
\usepackage{flushend}
\usepackage{color}
\usepackage{hyperref}
\usepackage{verbatim} 
\usepackage{wrapfig}
\usepackage{siunitx}

\begin{document}

\title{Density Encoding Enables Resource-Efficient Randomly Connected Neural Networks}


\author{Denis~Kleyko,
	Mansour~Kheffache,
        E.~Paxon~Frady,
        Urban~Wiklund,
        and~Evgeny~Osipov
\thanks{Manuscript received February 10, 2020; revised July 2, 2020; accepted August 8, 2020. 
This work  was  supported in part  by the Swedish Research Council (grant No. 2015-04677). 
The work of DK was supported by the European Union's Horizon 2020 Research and Innovation Programme under the Marie Skłodowska-Curie Individual Fellowship Grant Agreement 83917, and DARPA's VIP program under Super-HD project.
}
\thanks{\mbox{*}D. Kleyko is with the Redwood Center for Theoretical Neuroscience at the University of California, Berkeley, CA 94720, USA and also with Intelligent Systems Lab at Research Institutes of Sweden, 164 40 Kista, Sweden. \mbox{E-mail}: \mbox{denis.kleyko@ri.se}}
\thanks{M. Kheffache is with Netlight Consulting AB, 111 53 Stockholm, Sweden. \mbox{E-mail}: \mbox{mansour.kheffache@netlight.com}}
\thanks{E. P. Frady is with the Redwood Center for Theoretical Neuroscience at the University of California, Berkeley, CA 94720, USA. \mbox{E-mail}: \mbox{epaxon@berkeley.edu}
}
\thanks{U. Wiklund is with the Department of Radiation Sciences, Biomedical Engineering, Ume\r{a} University, 901 87 Ume\r{a}, Sweden. \mbox{E-mail}: \mbox{urban.wiklund@umu.se}}
 \thanks{E. Osipov is with the Department of Computer  Science Electrical and Space Engineering, Lule\aa{} University of Technology, 971 87 Lule\aa{}, Sweden. \mbox{E-mail}: \mbox{evgeny.osipov@ltu.se} }
 }

\markboth{IEEE TRANSACTIONS ON NEURAL NETWORKS AND LEARNING SYSTEMS}%
{Kleyko \MakeLowercase{\textit{et al.}}: Density Encoding Enables}

\maketitle

%
\begin{abstract}
The deployment of machine learning algorithms on resource-constrained edge devices is an important challenge from both theoretical and applied points of view.
In this article, we focus on resource-efficient randomly connected neural networks known as Random Vector Functional Link (RVFL) networks since their simple design and extremely fast training time make them very attractive for solving many applied classification tasks. 
We propose to represent input features via the density-based encoding  known in the area of stochastic computing and use the operations of binding and bundling from  the area of hyperdimensional computing for obtaining the activations of the hidden neurons. 
Using a collection of $121$ real-world datasets from the UCI Machine Learning Repository, we empirically show that the proposed approach demonstrates higher average accuracy than the conventional RVFL.
We also demonstrate that it is possible to represent the readout matrix using only integers in a limited range with minimal  loss in the accuracy. 
In this case, the proposed approach operates only on small \textit{n}-bits integers, which results in a computationally efficient architecture. 
Finally, through hardware  
Field-Programmable Gate Array (FPGA)
implementations, we show that such an approach consumes approximately eleven times less energy than that of the conventional RVFL.
\end{abstract}


\section{Introduction}
\label{sect:intro}

An ability to provide insights and predictive analytics in real-time is the greatest 
demand from businesses and industries to data-driven technologies.  
The vector of the current development targets enabling machine learning applications on connected 
devices (edge computing) such as smartphones, robots, vehicles, etc. 
The benefits of computing at the edge are tremendous: higher reliability of 
solutions due to the decoupling from the network connectivity and bandwidth availability; very low latency; higher security and privacy as sensitive data are processed locally on a device.

Randomly connected neural networks such as the recently proposed class of advanced randomized learning techniques called Stochastic Configuration Networks~\cite{SMC} and the well-known Random Vector Functional Link (RVFL)~\cite{RVFLorig} have become an increasingly popular topic of modern theoretical and applied research. 
On the theoretical side, the main result is that RFVLs provide universal approximation for continuous maps and functional approximations that converge in Kullback-Leibler divergence, when the target function is a probability density function~\cite{RandMM}. 
When this is combined with the simplicity of RVFL's design and training process, it makes them 
a very attractive alternative for solving practical machine learning problems in edge computing. 


The aim of this article is to present an approach for an order of magnitude increase of the 
resource-efficiency (memory footprint, computational complexity, and energy consumption)  
of RVFLs operations.
The proposed approach combines techniques from two fields of computer science: stochastic computing~\cite{ComputingRandomness} and hyperdimensional computing~\cite{Kanerva09}. 
The fundamental idea  is in the realization of activations of the hidden layer with the computationally simple operations 
of hyperdimensional computing, and the usage of the density-based encoding of the input features as in stochastic computing. 
Moreover, we enhance this approach with the integer-only readout matrix.
This combination allows us to use integer arithmetics end-to-end.
The novel contributions of the article are as follows:  
\begin{itemize} 

\item 
A resource-efficient approach to RVFLs is proposed, which uses only integer operations; 

\item The empirical evaluation on $121$ real-world classification datasets demonstrates that the accuracy of the proposed approach is higher than that of the conventional RVFL;

\item  
Field-Programmable Gate Array (FPGA)
implementation of the proposed approach is an order of magnitude more energy-efficient and $2.5$ times faster than the conventional RVFL.
\footnote{
For a network with: $16$ features, $4$ classes, and $512$ hidden neurons, which are the median values for the considered $121$ datasets.
}


\end{itemize}

The article is structured as follows. 
The background of methods used for the proposed approach is presented in Section \ref{sect:methods}. 
The approach itself is described in Section \ref{sect:density:RVFL}. 
The performance evaluation follows in Section \ref{sect:perf}. 
Section~\ref{sect:related} covers related work.
Section \ref{sect:conclusions} presents the concluding remarks.








\section{Background and methods}
\label{sect:methods}


\subsection{Random Vector Functional Link}
\label{sect:rvfl}

\begin{figure}[tb]
\centering
\includegraphics[width=0.7\columnwidth]{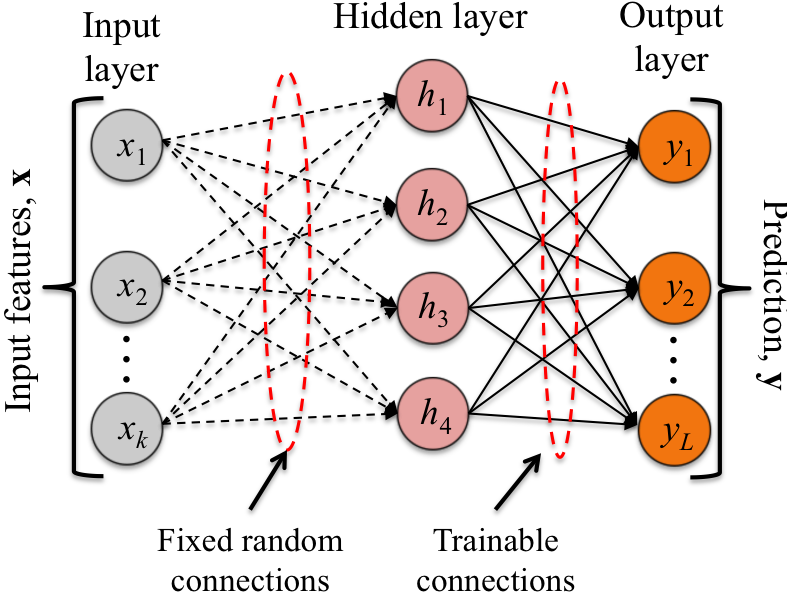}
\caption{The architecture of the conventional Random Vector Functional Link. In the presented example, the number of hidden neurons is set to $N=4$.} 
\label{fig:elm}
\end{figure}

This subsection briefly describes the conventional RVFL. 
For a detailed survey of RVFLs diligent readers are referred to~\cite{RCNNSsurvey}.  
Fig.~\ref{fig:elm} depicts the architecture of the conventional RVFL, which includes three layers of neurons. 
The input layer with $K$ neurons represents the current values of input features denoted as $\textbf{x}  \in [K \times 1]$. 
The output layer ($L$ neurons) produces the prediction of the network (denoted as $\textbf{y}$) during the operational phase. 
The layer in the middle is the hidden layer of the network, which performs a nonlinear transformation of input features.
The hidden layer contains $N$ neurons and its state is denoted as $\textbf{h}$.

In general, the connectivity of an RVFL is described by two matrices and a vector. 
A matrix $\textbf{W}^{\text{in}}  \in [N \times K]$ describes connections between the input layer neurons and the hidden layer neurons. 
This matrix projects the given input features to the hidden layer.
Each neuron in the hidden layer has a parameter called a bias. 
Biases of the hidden layer are stored in a vector and denoted as $\textbf{b} \in [N \times 1]$.   
The other matrix of readout connections $\textbf{W}^{\text{out}} \in [L \times N]$ between the hidden and the output layers transforms the current activations in the hidden layer stored in $\textbf{h}$
into the network's output $\textbf{y}$.\footnote{Strictly speaking, in the most general case, the readout matrix could also include connections between the input layer and the output layer. 
However, in the scope of this study, we only consider the case when the output layer predictions are obtained from the activations of the hidden layer.  
The interested readers are referred to work~\cite{RVFLconnections}, which performed a comprehensive evaluation of different design choices for the RVFL.
}

The main feature of the RVFL is that matrix $\textbf{W}^{\text{in}}$ and vector $\textbf{b}$ are randomly generated at the network initialization and stay fixed during the network's lifetime. 
There are no strict limitations for the generation of $\textbf{W}^{\text{in}}$ and $\textbf{b}$. 
They are usually randomly drawn from either normal or uniform distributions. 
Here, both $\textbf{W}^{\text{in}}$ and $\textbf{b}$ are generated from a uniform distribution. Following~\cite{RVFLFPGA}, the range for $\textbf{W}^{\text{in}}$ is $[-1, 1]$ while the range for $\textbf{b}$ is $[-0.1, 0.1]$.
Since $\textbf{W}^{\text{in}}$ and $\textbf{b}$ are fixed the process of training RVFL is focused on learning the values of the readout matrix $\textbf{W}^{\text{out}}$. 
The main advantage of training only $\textbf{W}^{\text{out}}$ is that the corresponding optimization problem is strictly convex, thus, the solution could be found in a single analytical step.

The activations of the network's hidden layer $\textbf{h}$ are described by the following equation:
\noindent
\begin{equation}
\textbf{h}=g ( \textbf{W}^{\text{in}}\textbf{x}+\textbf{b}),
\label{eq:rvfl:hid}
 \end{equation}
\noindent
where $g(x)$ is a nonlinear activation function applied to each neuron. 
Here, the sigmoid function $g(x)=\frac{1}{1+e^{-x}}$ is used.
Thus, the activation function restricts the range of possible activation values in the hidden layer to the range $[0, 1]$.

The predictions issued by the output layer are calculated as:  
\noindent
\begin{equation}
\textbf{y}= \textbf{W}^{\text{out}} \textbf{h}.
\label{eq:rvfl:y}
 \end{equation}

With respect to the training of RVFLs, the article focuses on classification tasks\footnote{Though,  the proposed approach is also applicable to regression problems. One may, however, expect that the quality of predictions might be more sensitive to the use of the density-based encoding.} considering only supervised-learning scenarios when the network is provided with the ground truth label for each training example. 
The total size of the training dataset is denoted as $M$.
In this setting, the standard way of acquiring weights of the trainable connections between the hidden and the output layers in  $\textbf{W}^{\text{out}}$ matrix is 
via solving the ridge regression (which is a special case of Tikhonov regularization) problem, 
which minimizes the mean square error between predictions (\ref{eq:rvfl:y}) and the ground truth. 
In particular, the activations of the hidden layer $\textbf{h}^T$ for each training example are collected together in matrix $\textbf{H}  \in [M \times N]$.   
Matrix $\textbf{Y}  \in [M \times L]$ stores the corresponding ground truth classifications using one-hot encodings.
Given $\textbf{H}$ and $\textbf{Y}$, $\textbf{W}^{\text{out}}$ is calculated as follows:
\noindent
\begin{equation}
\textbf{W}^{\text{out}} =  (\textbf{H}^T \textbf{H}+ \lambda \textbf{I})^{-1} \textbf{H}^T \textbf{Y},
\label{eq:rvfl:wout}
 \end{equation}
\noindent
where $\textbf{I}$ denotes an identity matrix of the suitable dimensionality ($\textbf{I}  \in [N \times N]$); $\lambda$ is a hyperparameter (scalar) determining the weight of the regularization part.\footnote{Note that (\ref{eq:rvfl:wout}) is computationally simpler compared to, e.g., the backprop algorithm and it is implementable efficiently on CPUs as well as on GPUs.}

\subsection{Density-based encoding of scalars}
\label{sect:density:encoding}



The idea of representing scalars as vectors is not new. 
It has been independently proposed in several areas. 
The area of stochastic computing~\cite{ComputingRandomness} is probably the most notable example since the whole idea of the stochastic computing is that it is possible to implement arithmetics on scalars using boolean operations on vectors (in general, streams) of bits. 
The rate coding model of neuronal firing used, e.g., in spiking neural networks is another notable example. 
Stochastic computing operates with scalars between $0$ and $1$, which are represented as random bit vectors where the scalar being encoded determines the probability of generating ones. 
Thus, the density of ones in the obtained bit vector encodes the scalar, hence such a representation method is called the density-based encoding. 
Generating random streams is important because the independence of two vectors is a prerequisite for using boolean operations to implement the arithmetics on them (e.g., AND for multiplication). 
Note that for the proposed approach no arithmetic operations will be performed with  the density-based encodings of scalars.
Therefore, the randomness of representations for encoding scalars is not compulsory in this study.
In fact, from the simplicity point of view, it is more advantageous to use a structured version of the density-based encoding, which does not require a source of randomness. 
We will use the structured version of the density-based encoding also known under the name thermometric encoding~\cite{Scalarencoding} for the rest of this paper. 

\begin{figure}[tb]
\centering
\includegraphics[width=0.5\columnwidth]{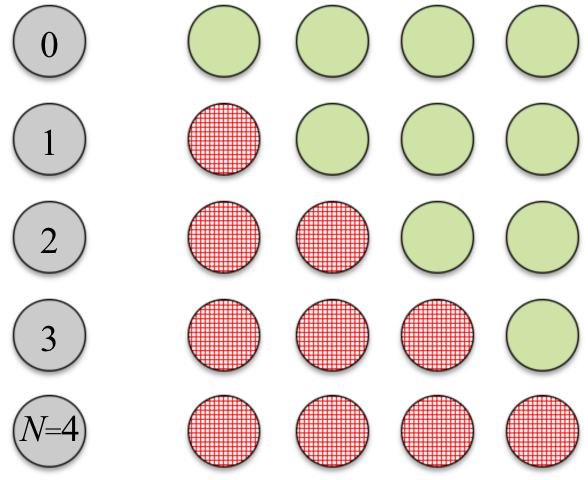}
\caption{An example of the density-based encoding when the dimensionality of representation is set to $N=4$.
}
\label{fig:repr:ex}
\end{figure}

The most intuitive way of presenting the concept of the density-based encoding is via visualization. 
Fig.~\ref{fig:repr:ex} illustrates all possible values, which could be encoded when dimensionality of the representation\footnote{It will become evident in Section~\ref{sect:density:RVFL} why the same notation $N$ as for the number of hidden neurons is used.} is set to $N=4$.
Fig.~\ref{fig:repr:ex} indicates that using the density-based encoding it is possible to represent $N+1$ different values. 
The most convenient way of denoting these values is by using integers in the range $[0, N]$ (nodes on the left in the figure). 
In this case, in order to obtain the encoding of a given value $v$, it is necessary to set $v$ leftmost positions of the vector to  ``one'' (hashed red nodes in the figure) while the rest of the vector is set to ``zero'' (filled green nodes). 
In the case of bipolar representations used below, ``one'' corresponds to $-1$ while ``zero'' corresponds to $1$.

Recall, however, that input features are not integers in the range $[0, N]$. 
Instead, it is assumed that a feature $x_i$ is represented by a real number in the range $[0,1]$.   
The task is to represent the current value of the feature as a vector $\textbf{f} \in [N \times 1]$ using the above density-based encoding. 
Since, the encoding requires a finite set of values between 0 and $N$, real numbers are first discretized using a fixed quantization step, which is determined by $N$. Given the current value of the feature, it is quantized to the closest integer as:
\noindent
\begin{equation} 
v=\lfloor x_i N \rceil, 
\label{eq:quantization} 
\end{equation}
\noindent
where $\lfloor * \rceil$ denotes rounding to the the closest integer. 
The obtained $v$ will determine the density-based encoding $\textbf{f}$. 
The presented procedure allows generating density-based encodings for the whole feature vector $\textbf{x}$.  
Matrix $\textbf{F} \in [N \times K]$, where $K$ denotes the number of features, contains the density-based encodings $\textbf{f}$ of the current values of $\textbf{x}$.

\subsection{Hyperdimensional computing}
\label{sect:sparse}

Hyperdimensional computing \cite{PlateBook,  Gallant} also known as Vector Symbolic Architectures is a family of bio-inspired methods of representing and manipulating concepts for cognitive architectures and their meanings in a high-dimensional space.
Vectors of high (but fixed) dimensionality (denoted as $N$) are the basis for representing information in hyperdimensional computing.\footnote{
These vectors are referred to as high-dimensional vectors or HD vectors.
}  
The information is distributed across HD vector's positions, therefore, HD vectors use distributed representations. 
Distributed representations~\cite{Hinton1986} are contrary to the localist representations since any subset of the positions can be interpreted. 
This is very relevant to the density-based encoding introduced in the previous subsection since the encoding in $\textbf{f}$ is also distributed.

In the scope of this article, columns of $\textbf{W}^{\text{in}}$  matrix are interpreted as HD vectors, which are generated randomly. 
These HD vectors are bipolar ($\textbf{W}^{\text{in}} \in \{-1, +1\}^{[N \times K]}$) and random with equal probabilities for $+1$ and $-1$. 
It is worth noting that an important property of high-dimensional spaces is that with an extremely high probability all random HD vectors are dissimilar to each other (quasi-orthogonal).
In order to manipulate HD vectors, hyperdimensional computing defines operations on them.
In this article, we implicitly use only two key operations: binding and bundling.

The binding operation is used to associate two HD vectors together. The result of binding is another HD vector. Here, the result of binding (denoted as $\textbf{z}$) two vectors $\textbf{x}$ and $\textbf{y}$  is calculated as follows: 
$\textbf{z} = \textbf{x}  \odot \textbf{y} $, 
where the notation $\odot$ for the Hadamard product is used to denote the binding operation since this article uses position-wise multiplication for binding. 
An important property of the binding operation is that the resultant HD vector $\textbf{z}$ is quasi-orthogonal to the HD vectors being bound.

The second operation is called bundling. 
The bundling operation combines several HD vectors into a single HD vector. 
Its simplest realization is a position-wise addition. 
However, when using the position-wise addition, the vector space becomes unlimited, therefore, it is practical to limit the values of the result.
This could be achieved with, e.g., a clipping function (denoted as $f_\kappa ( * )$):
\noindent
\begin{equation}
f_\kappa (x) = 
\begin{cases}
-\kappa & x \leq -\kappa \\
x & -\kappa < x < \kappa \\
\kappa & x \geq \kappa
\end{cases}
\label{eq:clipping}
\end{equation}
\noindent
In the clipping function, $\kappa$ is a configurable threshold parameter. 
Thus, in this article, the bundling operation is implemented via position-wise addition limited via the clipping function. 
For example, the result (denoted as $\textbf{a}$) of bundling HD vectors $\textbf{x}$ and $\textbf{y}$    is simply: 
$\textbf{a} = f_\kappa ( \textbf{x} + \textbf{y} )$.
In contrast to the binding operation, the resultant HD vector $\textbf{a}$ is similar to all bundled HD vectors, which allows, e.g., storing information in HD vectors~\cite{Frady17}.
For example, we have demonstrated the usefulness of the clipping function for resource-efficient implementations of Self-Organizing Maps~\cite{intSOM} and Echo State Networks~\cite{intESN, NepomnyashchiyHardwareESN2020}.

\section{RVFL with density-based encodings}
\label{sect:density:RVFL}

\begin{figure}[tb]
\centering
\includegraphics[width=0.9\columnwidth]{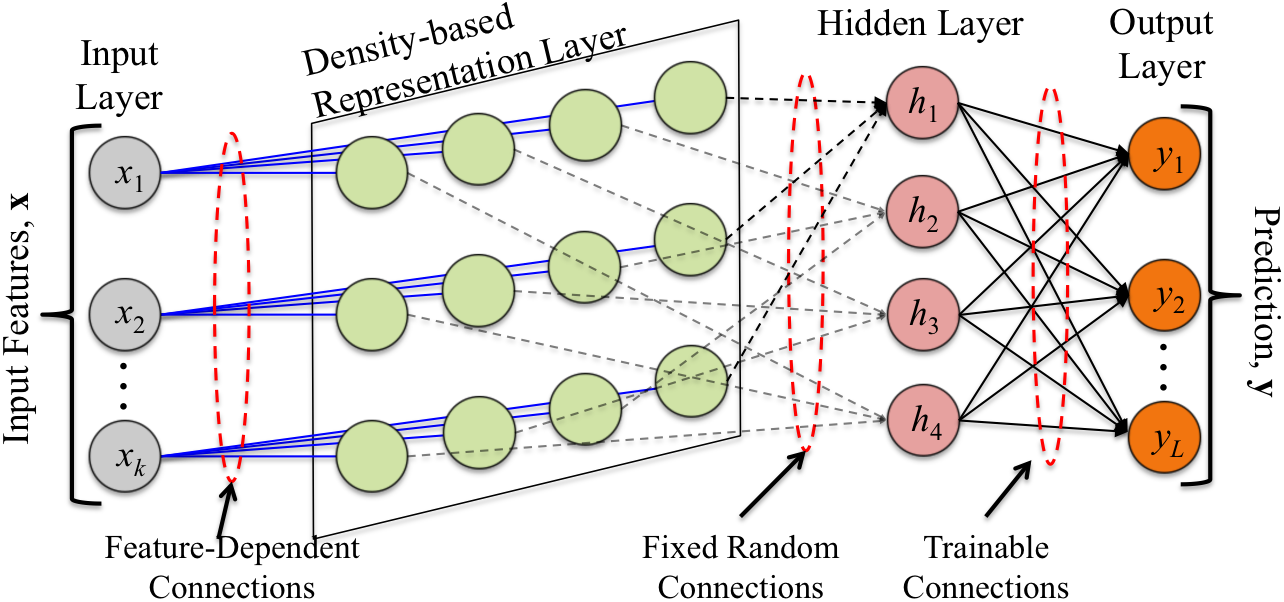}
\caption{The architecture of the Random Vector Functional Link, which relies on the density-based encoding.
In the presented example, the number of hidden neurons as well as the dimensionality of encoding are set to $N=4$.
}
\label{fig:intelm}
\end{figure}

\begin{figure*}[tb]
\centering
\includegraphics[width=2.0\columnwidth]{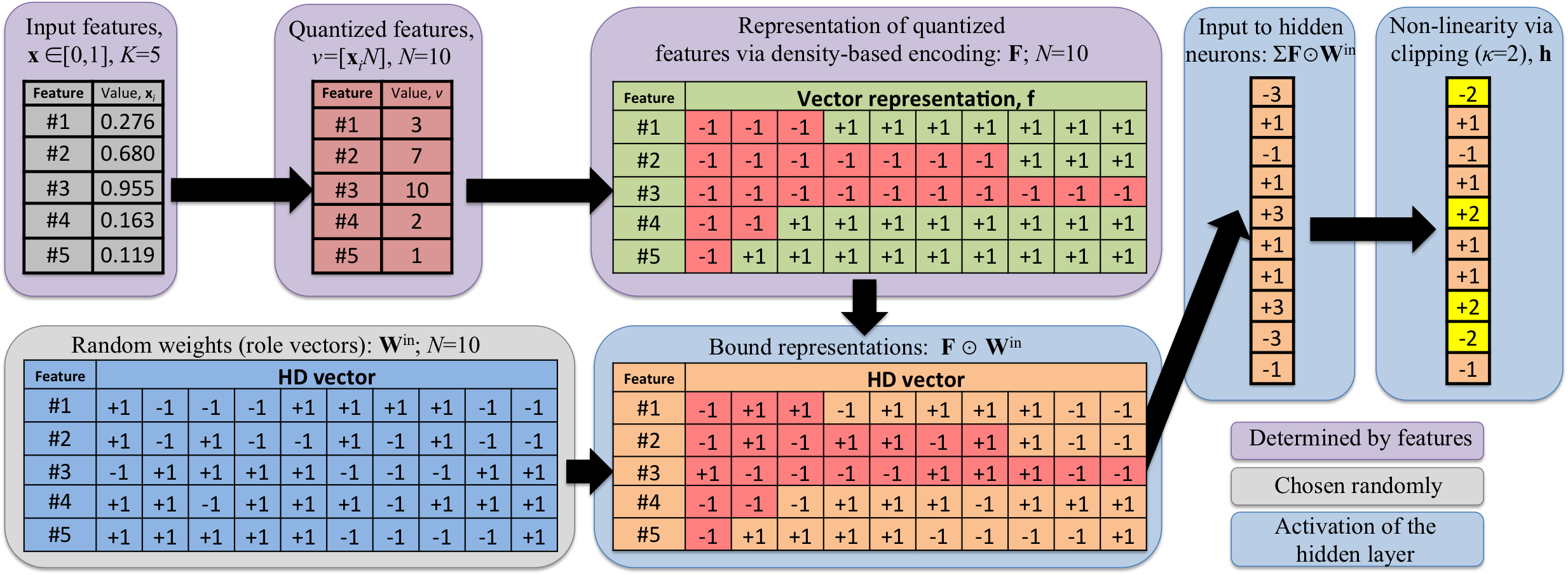}
\caption{
An example of activating the hidden layer with density-based encodings: $K=5$; $N=10$. Note that $N$ is set to $10$ for visualization purposes only.
}
\label{fig:intelm:ex}
\end{figure*}

This section presents an architecture of the RVFL utilizing the density-based encoding. 
The approach is illustrated in Fig.~\ref{fig:intelm}. 
The architecture is intentionally depicted to be as structurally identical to the conventional RVFL (Fig.~\ref{fig:elm}) as possible.
The major difference is that the proposed approach is illustrated with four layers of neurons: input layer ($\textbf{x}$, $K$ neurons); 
density-based representation layer ($\textbf{F}$, $N\times K$ neurons);
hidden layer ($\textbf{h}$, $N$ neurons); and
output layer ($\textbf{y}$, $L$ neurons).
Thus, in contrast to the conventional RVFL, the hidden layer is not connected directly to the input layer. 
Instead, each input feature is first transformed to a row of neurons storing its density-based encodings. 
These vectors constitute the density-based representation layer, which in turn is connected to the hidden layer. 
Note also, that the input and density-based representation layers are not fully-connected. 
Each neuron in the input layer is only connected to $N$ neurons in the corresponding row of the next layer. 
Moreover, these connections (blue lines in Fig.~\ref{fig:intelm}) are called  ``feature-dependent'' because the activation of the $i$-th input neuron $\textbf{x}_i$ will be quantized to the closest integer $v$ according to (\ref{eq:quantization}); in turn $v$ determines the number of the rightmost connections, which transmit $-1$, the remaining connections from that neuron transmit $+1$. 
Since each neuron in the density-based representation layer has only one incoming connection, the input activations are projected in the form of the bipolar matrix $\textbf{F}$.

It is also important to mention that the density-based representation and hidden layers are not fully-connected. 
In fact, each neuron in the density-based representation layer has only one outgoing connection.
Therefore, the matrix $\textbf{W}^{\text{in}}$ describing the fixed random connections to the hidden layer is still $\textbf{W}^{\text{in}}  \in [N \times K]$.
Moreover, these connections have a clear structure. 
In Fig.~\ref{fig:intelm} the connections are structured in such a way that each column in $\textbf{F}$ is connected to one of the hidden layer neurons. 
It explains why the number of hidden neurons $N$ also determines the dimensionality of the density-based encoding of features: each hidden neuron has its corresponding column in $\textbf{F}$ (see Fig.~\ref{fig:intelm:ex}). 
Note that in Fig.~\ref{fig:intelm:ex} $N$ is set to $10$ only for visualization purposes. 
In practice, the values of  $N$ are larger.

Similar to the conventional RVFL, the values of $\textbf{W}^{\text{in}}$ are also generated randomly. However, the values are drawn equiprobably from $\{ -1,+1\}$. 
Thus, similar to \textbf{F}, $\textbf{W}^{\text{in}}$ is also a bipolar matrix. 
When reflecting to the ideas  of hyperdimensional computing, $\textbf{W}^{\text{in}}$ should be interpreted as $K$ $N$-dimensional bipolar HD vectors. 
In other words, each feature is assigned with the corresponding HD vector. 
Thus, a conceptual intermediate step before getting input values of the hidden neurons is the binding operation between features' HD vectors and their current density-based encoding.

Finally, the proposed approach uses different nonlinear activation function in the hidden layer, the clipping function (\ref{eq:clipping}) is used instead of the sigmoid function.
The clipping function is characterized by the threshold value $\kappa$ regulating nonlinear behavior of the neurons and limiting the range of activation values. 
Summarizing aforementioned differences, activations of the hidden layer $\textbf{h}$ are obtained as follows:
\noindent
\begin{equation}
\textbf{h}= f_\kappa (\sum \textbf{F} \odot  \textbf{W}^{\text{in}}),
\label{eq:intrvfl:hid}
 \end{equation}
\noindent
where 
$\sum$ is a column-wise summation.  
Note that in contrast to (\ref{eq:rvfl:hid}) there is no bias term since it has been found empirically that its presence does not improve classification performance. 
Once the activations of the hidden layer $\textbf{h}$ are obtained, the rest of the network works in the same way as the conventional RVFL. The predictions in $\textbf{y}$ are calculated according to (\ref{eq:rvfl:y}).

In order to make operations of the proposed approach more intuitive, Fig.~\ref{fig:intelm:ex} presents a numerical example of acquiring the activations of the hidden layer.
First, the input layer with $K=5$ neurons sets the values of the current feature vector. 
These values are quantized to integers in the range [0,10] (since $N=10$). 
The quantized values determine the neurons of the density-based encoding, which are set to $-1$ (the rest is $+1$). 
For example, since the third feature is quantized to $v=10$ all values of its density-based encoding are set to $-1$. 
The bottom left panel shows a randomly generated $\textbf{W}^{\text{in}}$. 
Once $\textbf{F}$ is obtained, we calculate the Hadamard product $\textbf{F} \odot  \textbf{W}^{\text{in}}$, which is denoted  as ``bound representations'' in Fig.~\ref{fig:intelm:ex}.
The row-wise summation of the resultant matrix represents the input values of the hidden layer.
Finally, the clipping function ($\kappa=2$ in Fig.~\ref{fig:intelm:ex}) is used in the hidden layer to get $\textbf{h}$.

Note that due to the way of forming $\textbf{F}$ and  $\textbf{W}^{\text{in}}$, the input to the hidden layer neurons is always integers in the range $[-K, K]$.
Moreover, even after the clipping the activations of neurons are integers in the range $[-\kappa$ and $\kappa]$ (practically, $\kappa < K$). 
Thus, each hidden neuron can be represented using only $\lceil  \log_2(2\kappa+1) \rceil$ bits of memory. 
For example, when $\kappa=3$, there are seven unique activations of a neuron, which can be stored with just three bits. 
Last but not least, it is worth mentioning that for an efficient implementation the explicit calculation of $\textbf{F}$ is redundant.  
As it could be seen from Fig.~\ref{fig:intelm:ex}, the same result as $\textbf{F} \odot  \textbf{W}^{\text{in}}$ could be obtained if for each feature we use $v$ as an indicator of which signs should be changed in $\textbf{W}^{\text{in}}$.
As it will be shown in the next section, these properties give a major advantage over the conventional RFVL for resource-efficient implementation on digital hardware.


Since in the proposed approach the part of the network between the hidden and output layers is not modified, the simplest case is to train the readout matrix $\textbf{W}^{\text{out}}$ in the same manner as for the conventional RVFL (Section~\ref{sect:rvfl}). 
Note that while training the readout matrix $\textbf{W}^{\text{out}}$ according to (\ref{eq:rvfl:wout}) there is no need to normalize the activation values in $\textbf{h}$.
Moreover, since the goal is to obtain a very simplistic implementation, it is worth considering alternatives where $\textbf{W}^{\text{out}}$ would contain only integer values in a small limited range. 
In particular, we have considered three options: quantizing the result of regression (\ref{eq:rvfl:wout}); using a genetic algorithm (GA) initialized randomly; using GA initialized with the quantized result of regression. 
During the search, GA used the cost function for the generalized Learning Vector Quantization~\cite{GeneralizedLVQ}.

\section{Performance evaluation}
\label{sect:perf}

In this section, the proposed approach is verified in three scenarios.\footnote{
The diligent readers are kindly referred to the Supplementary materials, which provide additional  experimental  evaluation  to  further justify the proposed approach.
} 
The first scenario compares it against the conventional RVFL in the case when the weights of the readout matrix are real numbers for both approaches. 
The second scenario compares the results for the real-valued readout matrix against the considered strategies of obtaining integer-valued readout matrix.
The final scenario compares FPGA implementations of the proposed approach and the finite precision RVFL~\cite{RVFLFPGA} in the case of a limited energy budget. 
All reported results\footnote{
The implementation of the experiments reported in the article is available online via 
\url{https://github.com/denkle/Density-Encoding-Enables-Resource-Efficient-Randomly-Connected-Neural-Networks}.}
are based on $121$ real-world classification datasets obtained from the UCI Machine Learning Repository\footnote{Available online: \url{http://persoal.citius.usc.es/manuel.fernandez.delgado/papers/jmlr/data.tar.gz}.}~\cite{Dua:2019}. 
The considered collection of datasets has been initially analyzed in a large-scale comparison study of different classifiers and the interested readers are kindly referred to the original work~\cite{HundredsClassifiers} for more details.
The only preprocessing step was to normalize features in the range $[0, 1]$.
Finally, the reported accuracies were averaged across five independent initializations.

\begin{figure*}[tb]
\begin{center}
\begin{minipage}[h]{0.53\columnwidth}
\centering
\includegraphics[width=1.0\columnwidth]{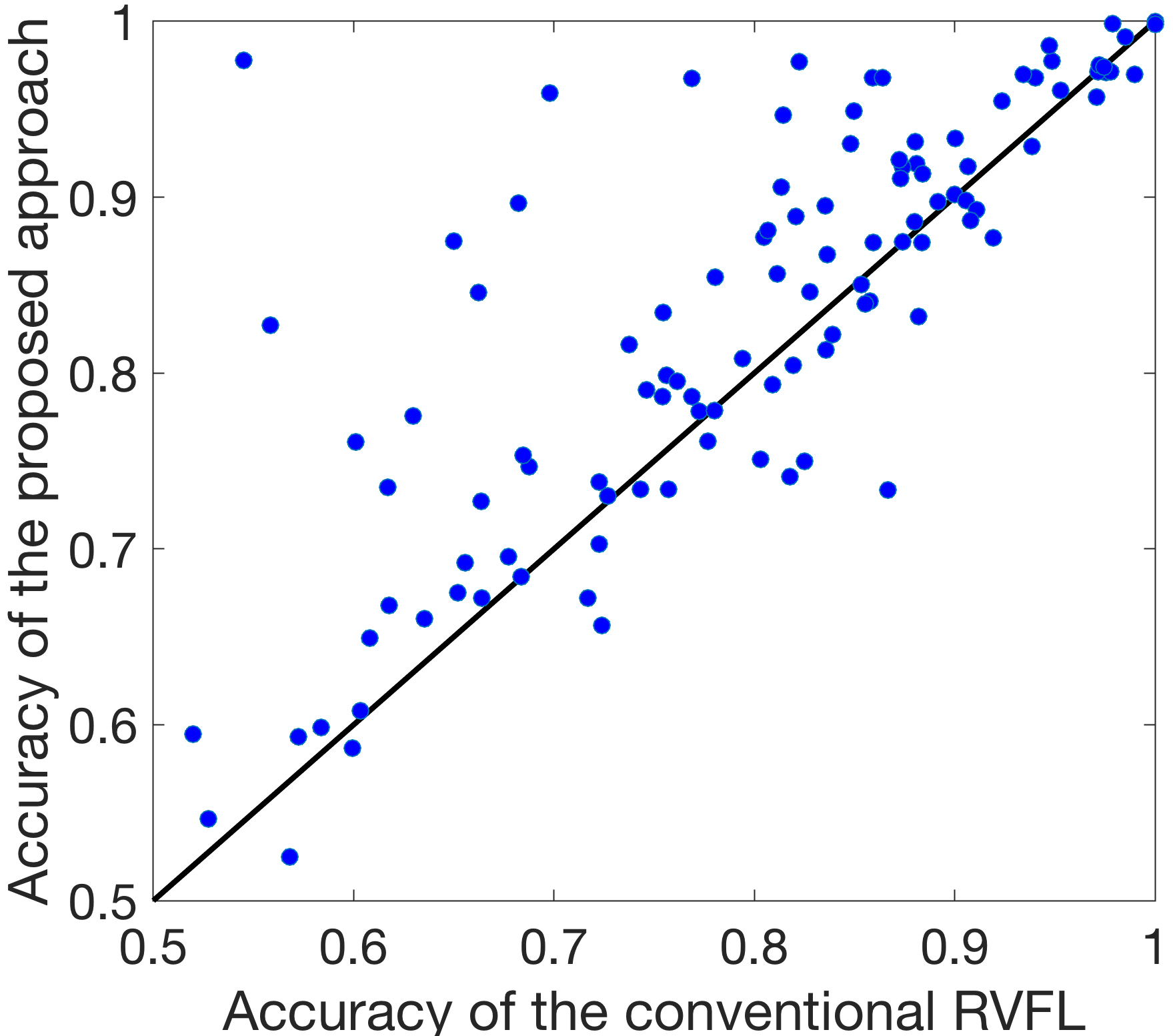}
\caption{
Cross-validation accuracy of the conventional RVFL against the proposed approach. 
A point corresponds to a dataset.
}
\label{fig:perf:sc1}
\end{minipage}
\hfill
\begin{minipage}[h]{0.85\columnwidth}
\includegraphics[width=1.0\columnwidth]{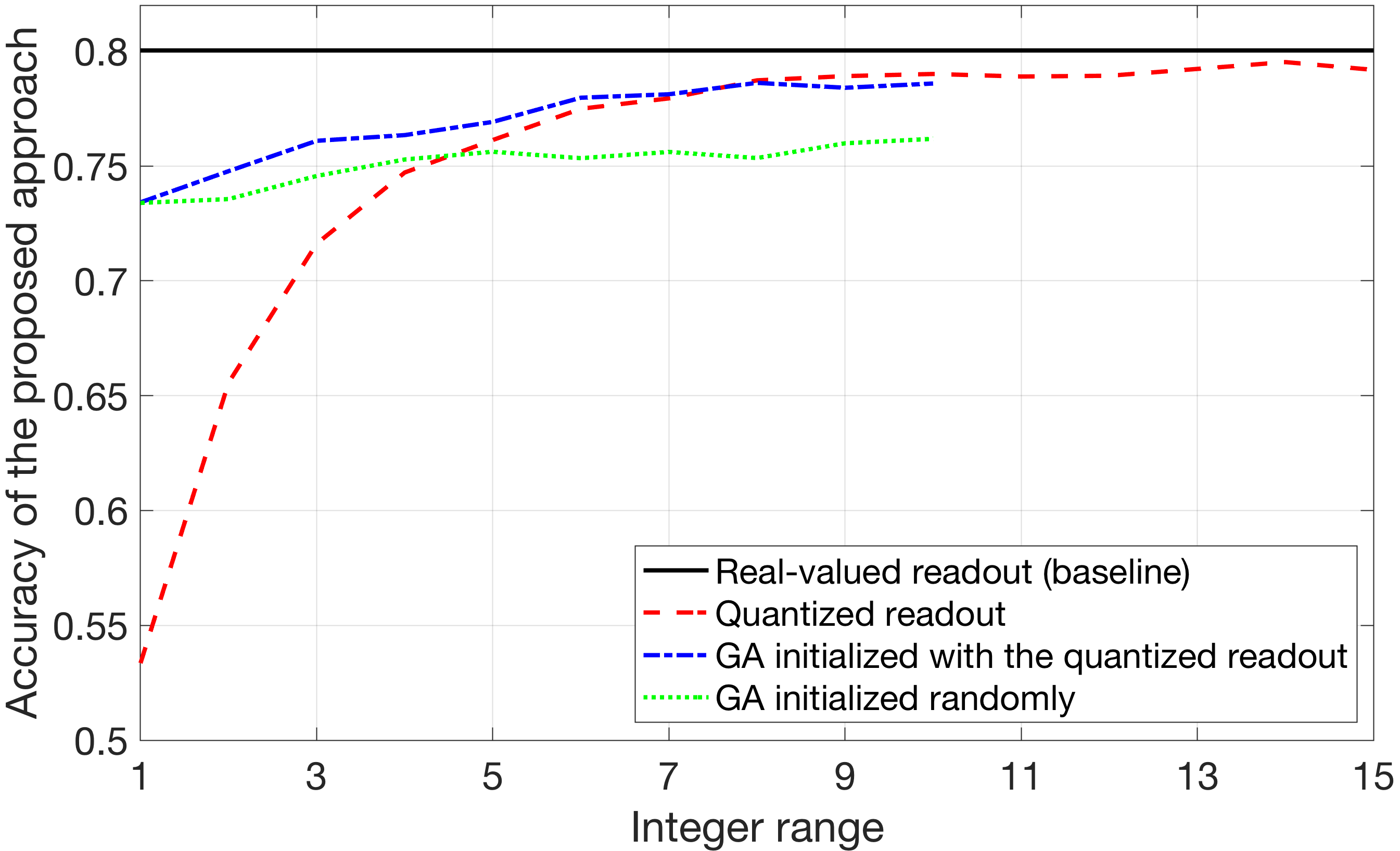}
\caption{
Average cross-validation accuracy of the proposed approach for different integer readout strategies. 
}
\label{fig:perf:sc2}
\end{minipage}
\hfill 
\begin{minipage}[h]{0.53\columnwidth}
\includegraphics[width=1.0\columnwidth]{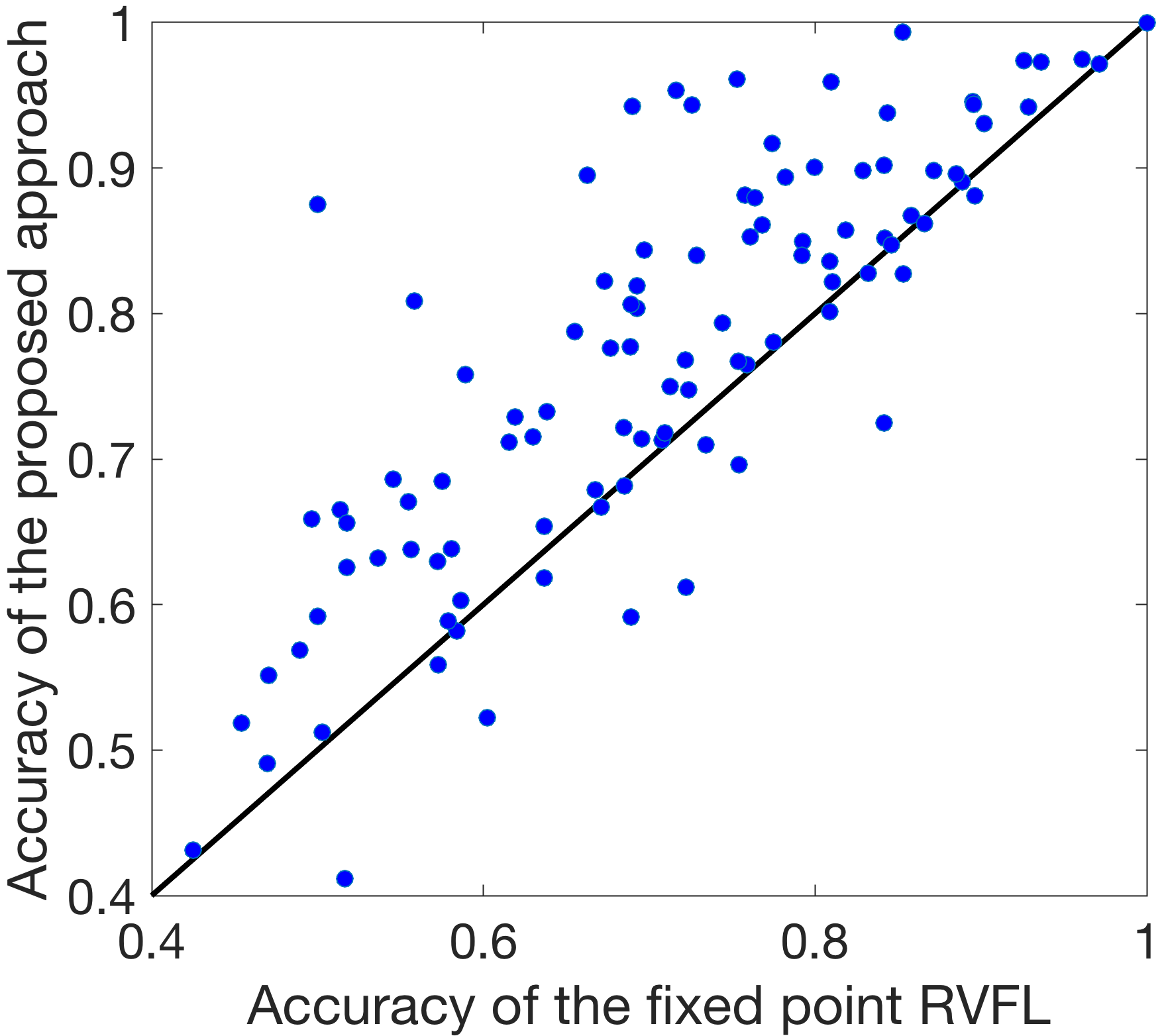}
\caption{
Cross-validation accuracy of the finite precision RVFL against the proposed approach in case of the fixed energy budget.
}
\label{fig:perf:sc3}
\end{minipage}
\end{center}
\end{figure*}

\subsection{Comparison with the conventional RVFL}


First, we compare the conventional RVFL with the proposed approach when computational resources for both approaches are not limited. 
The search of the hyperparameters has been done according to~\cite{HundredsClassifiers} using the grid search over $\lambda$ and $N$ in the case of the conventional RVFL and additionally considering $\kappa$ for the proposed approach; $N$ varied in the range $[50, 1500]$ with step $50$;  $\lambda$ varied in the range $2^{[-10, 5]}$ with step $1$; and $\kappa$ varied between $\{1,3,7,15\}$.
The obtained optimal hyperparameters were used to estimate the cross-validation accuracy on all datasets. 
In order to visualize the results, we rely on the same approach as reported in~\cite{DNNSmallData}.
Fig.~\ref{fig:perf:sc1} presents the accuracy of the conventional RVFL against the proposed approach. 
First, it is important to note that as expected the correlation coefficient  between the obtained results is high ($0.86$). 
Moreover, the average accuracy for the conventional RVFL is $0.76$ while that for the proposed approach is $0.80$.\footnote{In~\cite{HundredsClassifiers} the highest mean accuracy $0.82$ was obtained for Random Forest.}
The difference in accuracy was statistically significant using 5\% significance level according to two-sample hypothesis testing.
It is not absolutely intuitive why the proposed approach demonstrates higher accuracy. 
Nevertheless, one hypothesis is that the quantization for the density-based encoding might provide extra regularization.\footnote{For the conventional RVFL with quantized inputs the mean accuracy was $0.753$ ($0.755$ for non-quantized). 
The correlation coefficient was $0.986$. 
Thus, the improvement cannot be caused barely by the input quantization.
Also, for the case when the conventional RVFL additionally used the connections between the input layer and the output layer, there was no significant improvement in the accuracy as the mean accuracy was $0.762$.
}

\subsection{The effect of quantized readout weights}

Fig.~\ref{fig:perf:sc2} presents the average accuracy of the proposed approach for three considered strategies of obtaining the readout matrix with integer values against the average accuracy from the previous experiment. 
The considered ranges are symmetric and the figure indicates only positive boundaries.
It is clear that if the result of regression is quantized (dashed line) to very few levels the accuracy is affected significantly. 
However, with the increased number of levels, the accuracy approaches the baseline and it is concluded that 5-bits per weight results in a very close approximation. 
Refining the quantized result of regression with GA (dash-dot line) certainly improved the accuracy for a small number of quantization levels, which is in line with the results in~\cite{RVFLFPGA}. However, using GA for the number of levels larger than six was not beneficial.
Random GA initialization (dotted line) decreased the accuracy.

\subsection{Performance in the case of limited resources}

The third experiment compares FPGA hardware implementations of the proposed approach and the finite precision RVFL in the case of a fixed energy budget per one classification pass. 
The idea of restricting the energy budget could be seen as an intuitive set-up for comparing bounded-optimality~\cite{BO95} of two approaches. 
Finite precision RVFL~\cite{RVFLFPGA} with $8$-bits per neuron/weight was used since it is more efficient than the conventional RVFL.\footnote{
When comparing FPGA implementations of the conventional RVFL and the proposed approach (integer readout) for a network where $K=16$, $N=512$, and $L=4$ (median values for the $121$ UCI datasets) the proposed approach consumed about $11$ times less energy and was $2.5$ times faster. 
} 
Following the conclusions from the previous experiment, the resolution of the readout weights of the proposed approach was set to $5$-bits.
Both approaches were deployed on ZedBoard FPGA and the energy consumption was estimated with the Xilinx Power Estimator tool.
The energy budget was set to \SI{3.2}{\micro\joule} to reflect a network with typical parameters.\footnote{Since in each dataset input and output sizes are fixed, the budget was enforced by determining number of hidden layer neurons within the budget.}
Fig.~\ref{fig:perf:sc3} presents the accuracy of the proposed approach (average $0.73$) against the finite precision RVFL (average $0.65$).
The difference in accuracy was statistically significant using 5\% significance level according to two-sample hypothesis testing.
Due to the limited resources, values are lower than in the first experiment, nevertheless, the results are impressive when the performance of our approach is compared to the fixed point RVFL.

\section{Related work}
\label{sect:related}

This section briefly describes the related work. First of all, the readers generally interested in neural networks, which rely on randomly created connections, are kindly referred to the survey in~\cite{RCNNSsurvey}.

\subsection{Paradigms used for the proposed approach}

In order to design a resource-efficient RVFL algorithm, the proposed approach combines the ideas from two areas.
These are the density-based encoding from stochastic computing and the binding and bundling operations from hyperdimensional computing. 
Since both are research fields on their own, here we only indicate the introductory papers facilitating entrance to the areas. 
The recent magazine article~\cite{ComputingRandomness} is probably the most approachable reading for stochastic computing.
With respect to hyperdimensional computing, the best starting point is the tutorial-like article~\cite{Kanerva09} by Kanerva.

\subsection{Resource-efficient RVFL}

Recall that even the conventional RVFL networks are considered to be one of the simplest approaches for machine learning.
This fact explains why the efforts on pushing the resource-efficiency of RVFL networks to the extreme are rather limited. 
The most relevant works in this direction are~\cite{RVFLFPGA, RVFLFPGAtrain}. 
Similar to the present study, both works use FPGA for hardware experiments. 
Moreover, both works rely on finite precision implementation for improving the resource-efficiency. 
The work~\cite{RVFLFPGAtrain}, however, heavily focuses on the process of obtaining the weights of the readout matrix, which is not the case here.    
The work~\cite{RVFLFPGA}, which focuses on the operational phase, is used here as the baseline for comparison with the proposed approach. 
However, none of the previous works in the area of RVFL, to the best of our knowledge, have been focusing on using the combination of the density-based encoding with the binding operation. 
As an important topic for future research, we see the theoretical characterization of the classification performance improvement obtained with the proposed approach.
As indicated earlier, one hypothesis is that quantization and the density-based encoding provide extra regularization. 
In order to move in this direction, the relevant works are related to a phenomenon of network generalization improvement, e.g., via adding a noise~\cite{Bishop95}  or discretizing quantitative features~\cite{Zaidi2017}.

\subsection{Simplification of neural networks}

Finally, it is worth mentioning that in recent years the simplification of computing architectures for neural networks is an important research topic. 
Notable examples are works \cite{GoogleIntNN, BinNN, QuanNN}, which have been evaluated on convolutional neural networks;
work~\cite{TernaryNNs} that has introduced networks with ternary activations and work~\cite{BitwiseNNs} that has introduced networks where all parameters are binary. 
It is worth mentioning that in contrast to the bit-wise networks~\cite{BitwiseNNs}, the proposed use of the density-based encoding does not require the binarization of the input features, which often worsens the accuracy.

\section{Conclusions}
\label{sect:conclusions}

This article proposed a resource-efficient fully-integer approach to randomly connected neural networks. 
The key enabler for efficiently obtaining activations of the hidden neurons is the combination of the representation of input features via the density-based encoding used in the stochastic computing, and the use of binding and bundling operations from hyperdimensional computing area. 
Integer values of the readout matrix could be obtained with minimal  loss in the accuracy, e.g., by simple rounding of the ridge regression solution, which in turn could be fine-tuned by the genetic algorithm. 
The empirical evaluation was performed on $121$ real-world datasets. 
The proposed approach demonstrated a higher average accuracy than the conventional RVFL networks while being $2.5$ times faster and consuming eleven times less energy (typical network on FPGA).
Finally, the accuracy of the proposed approach significantly prevailed that of the finite precision RVFL networks when both networks implemented on hardware were constrained to a fixed energy budget. 

Despite that this work has focused only on classification tasks it is worth mentioning that the proposed approach of forming the activations of the hidden layer should be seen as a generic structured representation scheme based on high-dimensional random projections that allow for direct learning of complex nonlinear functions. Therefore, a promising direction for future work is to develop an analytical theory similar to the capacity theory of such representations~\cite{Frady17} that would relate the quality of approximations based on the complexity of a nonlinear function, number of its inputs and outputs, and a number hidden neurons and their resolution.

Last but not least, we conjecture that the density-based encoding will be useful for developing resource-efficient versions of other neural networks. 
For example, as it has been recently shown in~\cite{GeometryBCNN} that binarizing initial layers of convolutional neural networks could easily harm their accuracy.
We expect that the density-based encoding of input features will solve this issue.


\bibliography{bica_short}

\end{document}